\documentclass[10pt,twocolumn,letterpaper]{article}

\usepackage[pagenumbers]{iccv} 

\newcommand{\ours}{\textbf{SCORE}\xspace}
\newlength\savewidth\newcommand\shline{\noalign{\global\savewidth\arrayrulewidth
  \global\arrayrulewidth 1pt}\hline\noalign{\global\arrayrulewidth\savewidth}}

\usepackage[accsupp]{axessibility}  
\usepackage{colortbl}
\usepackage{xcolor}  
\usepackage{color}
\usepackage{url}            
\usepackage{booktabs}       
\usepackage{amsfonts}       
\usepackage{nicefrac}       
\usepackage{microtype}      

\usepackage{graphicx}
\usepackage{comment}
\usepackage{xspace}
\usepackage{caption}
\usepackage{mathtools}
\usepackage{graphicx}
\usepackage{soul}
\usepackage{xspace}
\usepackage{enumitem}

\usepackage{bbding}
\usepackage{wasysym}
\usepackage{amsmath}
\usepackage{amssymb}
\usepackage{bbm}
\usepackage{bm}
\usepackage{float}
\usepackage{dblfloatfix}
\usepackage{footnote}
\usepackage{array} 
\usepackage{dblfloatfix}
\usepackage{transparent}
\usepackage{xspace}
\usepackage{multirow}
\usepackage{amsfonts}
\usepackage{pifont}
\usepackage{algorithm}
\usepackage{listings}
\usepackage{bbold}
\usepackage{wrapfig}

\usepackage{adjustbox}

\newcommand{\cmark}{\ding{51}\xspace}%
\newcommand{\xmarkg}{\textcolor{lightgray}{\ding{55}}\xspace}%
\definecolor{pptblue}{RGB}{66, 133, 244}
\definecolor{pptyellow}{RGB}{254,240,0}
\definecolor{pptorange}{RGB}{255,171,64}

\definecolor{iccvblue}{rgb}{0.21,0.49,0.74}
\usepackage[pagebackref,breaklinks,colorlinks,allcolors=iccvblue]{hyperref}

\def\paperID{3427} 
\def\confName{ICCV}
\def\confYear{2025}

\title{SCORE: \underline{S}cene \underline{C}ontext Matters \\in \underline{O}pen-Vocabulary \underline{Re}mote Sensing Instance Segmentation}

\author{
Shiqi Huang$^1$
\quad
Shuting He$^2$
\quad
Huaiyuan Qin$^3$
\quad
Bihan Wen$^1$\thanks{Corresponding author}\\
$^1$Nanyang Technological University\\
\quad
$^2$MoE Key Laboratory of Interdisciplinary Research of Computation and Economics, \\
Shanghai University of Finance and Economics\\
\quad
$^3$Institute for Infocomm Research (I\textsuperscript{2}R), A*STAR, Singapore
}

\begin{document}
\maketitle
\begin{abstract}
Most existing remote sensing instance segmentation approaches are designed for close-vocabulary prediction, limiting their ability to recognize novel categories or generalize across datasets. This restricts their applicability in diverse Earth observation scenarios. To address this, we introduce open-vocabulary (OV) learning for remote sensing instance segmentation. While current OV segmentation models perform well on natural image datasets, their direct application to remote sensing faces challenges such as diverse landscapes, seasonal variations, and the presence of small or ambiguous objects in aerial imagery. To overcome these challenges, we propose \ours (\textbf{S}cene \textbf{C}ontext matters in \textbf{O}pen-vocabulary \textbf{RE}mote sensing instance segmentation), a framework that integrates multi-granularity scene context, i.e., regional context and global context, to enhance both visual and textual representations. Specifically, we introduce Region-Aware Integration, which refines class embeddings with regional context to improve object distinguishability. Additionally, we propose Global Context Adaptation, which enriches naive text embeddings with remote sensing global context, creating a more adaptable and expressive linguistic latent space for the classifier. We establish new benchmarks for OV remote sensing instance segmentation across diverse datasets. Experimental results demonstrate that, our proposed method achieves SOTA performance, which provides a robust solution for large-scale, real-world geospatial analysis. Our code is available at \href{https://github.com/HuangShiqi128/SCORE}{https://github.com/HuangShiqi128/SCORE}.
\end{abstract}    
\section{Introduction}
\label{sec:intro}
Instance segmentation is a fundamental task in remote sensing, aiming to localize aerial objects with pixel-wise instance masks and category labels \cite{su2019object, su2020hq,  carvalho2020instance, zhang2021semantic, xu2021improved, liu2024learning, ye2023remote, cao2024obbinst}. It plays a vital role in diverse applications, including environmental monitoring \cite{guan2022forest, sani2022instance}, urban development \cite{wu2020improved, yasir2023instance, wei2022lfg, liu2020multiscale}, and agricultural planning \cite{zhu2024cug_misdataset, zhong2023multi}. With the vast amount of remote sensing data collected from satellites, drones, and aerial surveys, instance segmentation serves as a key tool for large-scale geospatial analysis.

\begin{figure}[t]
  \centering
   \includegraphics[width=0.95\linewidth]{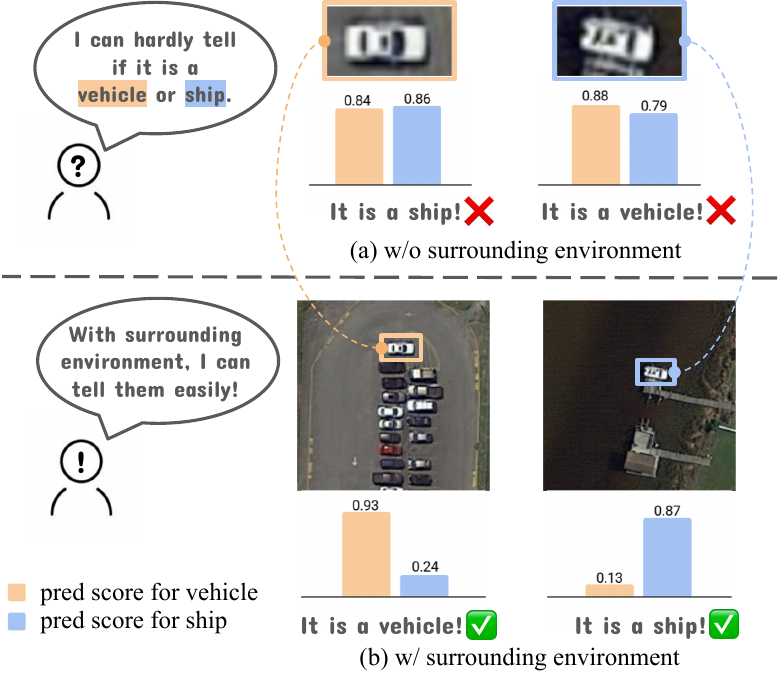}
   \caption{The example illustrates how human perception \& model prediction differ (a) without and (b) with the surrounding environment. Prediction scores are derived from FC-CLIP \cite{yu2023convolutions}.} 
   \vspace{-3mm}
   \label{fig:teaser}
\end{figure}

Most existing approaches rely on pixel-wise annotated datasets to train instance segmentation models \cite{su2019object, su2020hq, zamir2019isaid}. However, manually labeling aerial imagery is both time-consuming and challenging due to the small, densely packed, and visually ambiguous nature of instance objects in remote sensing \cite{yao2016semantic, hanyu2024aerialformer}. Furthermore, these models are constrained by the training data, limiting their ability to recognize novel categories or generalize across different domains. This restricts their applicability in diverse Earth observation tasks. Open-vocabulary (OV) learning \cite{wu2024towards, xu2022simple, zhang2023simple} offers a promising solution by enabling models to predict novel classes without requiring exhaustive annotations or retraining. This capability is particularly valuable for remote sensing, where  rapid adaptation is essential for monitoring evolving environmental conditions.

Although numerous OV segmentation models have been developed for natural images \cite{ding2023open, xu2023open, yu2023convolutions, jiao2024collaborative}, directly applying the models to remote sensing is less effective. This is due to the unique challenges posed by diverse landscapes, seasonal variations, and different imaging conditions that must be accounted for aerial and satellite imagery \cite{ye2025GSNet,cao2024open}. Additionally, aerial images, typically captured from a bird’s-eye view at large scales, often contain small or ambiguous objects that are difficult to recognize \cite{hanyu2024aerialformer, yao2016semantic}. For example, as shown in Figure \ref{fig:teaser}(a), differentiating between vehicles and ships can be challenging due to their shared elongated shapes and similar appearances. 

On the other hand, in remote sensing, objects are often closely correlated with their surrounding environment, providing useful prior knowledge \cite{li2023large, li2021robust, marcu2016dual}. For instance, ships are typically found near coastal areas, cars are associated with roads or parking lots, airplanes are near airports, and agricultural fields are common in rural areas. 
This regional scene context can help differentiate instance-level objects.
As illustrated in Figure \ref{fig:teaser}(b), given the presence of water and harbors, the object on the left is likely a ship, whereas the object on the right, situated in a parking lot, is a car.
Motivated by this, we propose leveraging \textbf{regional context}, derived from the surrounding environment, to enhance object representations. With the rise of remote sensing-specific vision-language models, \ie, remote sensing CLIPs \cite{liu2024remoteclip, zhang2023rs5m, wang2024skyscript, pang2024h2rsvlm}, 
these models encode valuable domain knowledge, enabling the extraction of meaningful regional scene context.
By utilizing regional context as a guiding cue, our approach aims to improve object recognition in remote sensing imagery, particularly in the challenging open-vocabulary settings.

While incorporating prior knowledge enhances recognition by associating instances with their regional context, existing OV segmentation models, which rely on frozen classifiers trained on natural images, lack domain-specific adaptability for remote sensing. 
Their fixed text embeddings struggle to capture the significant intra-class variance, resolution differences, and environmental diversity present in remote sensing datasets \cite{zhang2020well, huang2024zori, rsiscmdl}.
To address this challenge, we propose enhancing the text embeddings by incorporating domain-specific \textbf{global context}, a subset of scene context extracted from remote sensing CLIP. 
This adaptation enriches the classifier’s linguistic latent space with domain-relevant visual cues, ultimately improving class predictions for aerial objects.

Our main contributions can be summarized as follows:
\begin{itemize}
   \item We put forward open-vocabulary remote sensing instance segmentation task and develop a comprehensive framework \ours to improve OV instance segmentation performance across diverse remote sensing datasets. We establish new experimental benchmarks and achieve SOTA performance on the proposed task.
   \item We design Region-Aware Integration, leveraging the correlation between objects and their surrounding environment in aerial images. By incorporating regional context retrieved from domain-specific CLIP, our method enhances class embeddings, improving object classification in open-vocabulary remote sensing instance segmentation.
   \item To bridge the gap between general-domain and remote sensing-specific classifiers, we introduce Global Context Adaptation which injects aerial global context into CLIP text embeddings. This adaptation enhances richness of text representation, making them more expressive and suited for classifying remote sensing objects.
\end{itemize}

\section{Related Work}
\label{sec:related_work}

\begin{figure*}[t]
  \centering
    \includegraphics[width=0.95\textwidth]{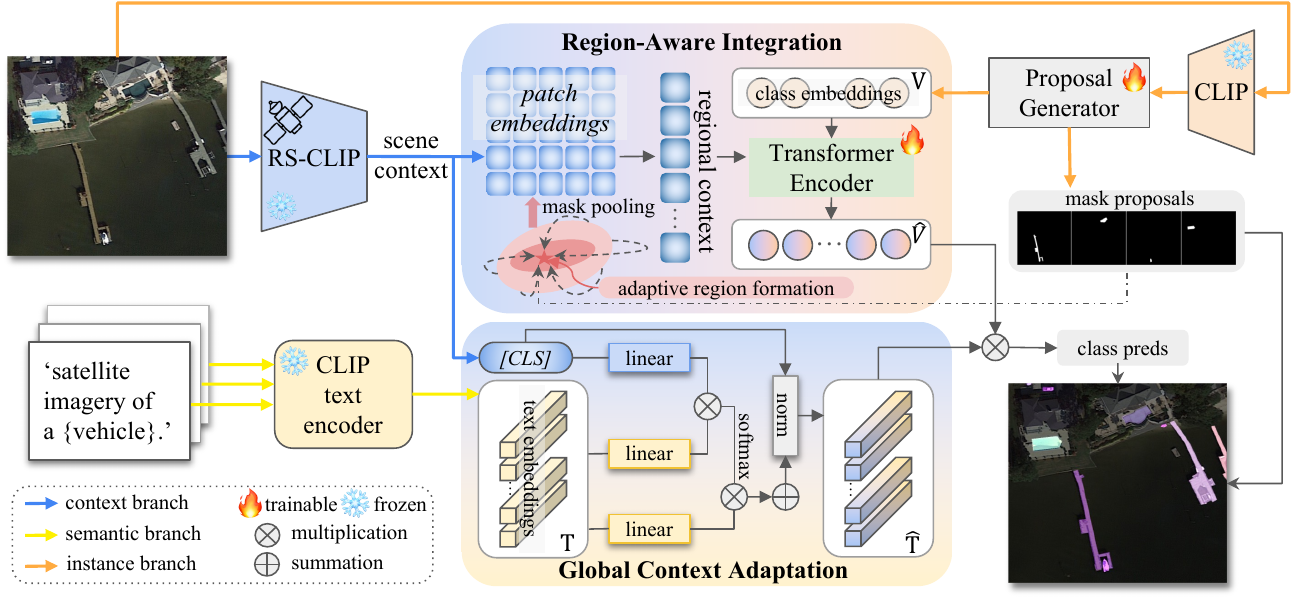}
   \caption{\textbf{Overview of \ours training framework.} It consists of three main branches: context branch ({\color{pptblue}blue}) extracts multi-granularity scene context from RS CLIP, semantic branch ({\color{pptyellow}yellow}) encodes text embeddings with a frozen CLIP text encoder, and instance branch ({\color{pptorange}orange}) generates class embeddings and instance proposals. These branches interact through Region-Aware Integration (RAI) and Global Context Adaptation (GCA) to derive scene-context-enhanced class embeddings $\hat{\mathbf{V}}$ and text embeddings $\hat{\mathbf{T}}$, which are used for class prediction. \textit{[CLS]} represents the domain-specific [CLS] token, \ie, global context.}
   \label{fig:framework}
   \vspace{-9pt}
\end{figure*}

\subsection{Remote Sensing Instance Segmentation}
With the advent of deep learning, segmentation frameworks for remote sensing images have made notable progress. For remote sensing instance segmentation, methods have been developed to tackle challenges such as scale variations, foreground-background confusion, and class ambiguity \cite{liu2024learning, ye2023remote, su2022faster, su2020hq, su2019object, liu2020global, zhang2021semantic, hanyu2024aerialformer}. More recently, the integration of foundation models into remote sensing segmentation has been explored. For example, RSPrompter \cite{chen2024rsprompter} enhances the instance segmentation capabilities of SAM \cite{kirillov2023segment} on remote sensing images using prompt learning. SAMRS \cite{wang2023samrs} expands remote sensing segmentation datasets by leveraging SAM. ZoRI \cite{huang2024zori} introduces zero-shot learning for remote sensing instance segmentation. Despite these advancements, the ability to make cross-dataset prediction, \ie, open vocabulary learning, for remote sensing instance segmentation has not been studied. 

\subsection{Open-Vocabulary Segmentation}
With the emergence of large-scale pre-trained multimodal foundation models like CLIP \cite{radford2021learning},  numerous methods have been developed for OV segmentation in natural images by leveraging their strong generalization capabilities \cite{xu2022simple,ding2023open,xu2023open, qin2023freeseg,chen2023open,yu2023convolutions,jiao2024collaborative,cho2024cat,xu2023side,liang2023open,li2022language,ghiasi2022scaling}. MaskCLIP \cite{ding2023open} adapts CLIP visual encoder with relative mask attention for universal OV segmentation. ODISE \cite{xu2023open} leverages the semantically differentiated representation pretrained in text-image diffusion models to generalize to novel classes. FreeSeg \cite{qin2023freeseg} builds a unified model that captures task-aware and category-sensitive concepts for more robust segmentation. OPSNet \cite{chen2023open} dynamically modulates query embeddings with CLIP features to enhance adaptation to novel ccategories. FC-CLIP \cite{yu2023convolutions}  incorporates a frozen CNN-based CLIP image encoder as the backbone, preserving cross-modal alignment while avoiding redundant feature extraction. MAFT+ \cite{jiao2024collaborative} further optimizes CLIP’s vision-text representations to improve OV segmentation results.

While significant progress has been made in OV segmentation for natural images, there is also a growing interest in extending it to the field of remote sensing. Recent works have explored OV remote sensing semantic segmentation \cite{ye2025GSNet,cao2024open}. OVRS \cite{cao2024open} introduces orientation-adaptive semantics to address the varying orientations of aerial objects. GSNet \cite{ye2025GSNet}
fuses features from CLIP backbone with remote sensing backbone to derive domain-specific representation.  However, existing methods are limited to semantic segmentation, there remains a critical need for an OV instance segmentation framework to address the challenges in remote sensing.

\subsection{Vision-Language Models}
\label{sec:related_work_vlm}
Vision-language models (VLMs) are designed to learn a shared representation space that bridges visual and textual modalities through contrastive learning. By pretraining on large-scale web-based datasets \cite{radford2021learning, jia2021scaling}, these models align image and text embeddings to capture meaningful semantic relationships. This cross-modal alignment enables VLMs to generalize effectively across a wide range of downstream tasks, including image retrieval, dense prediction, and cross-modal reasoning \cite{liu2021image, rao2022denseclip, gu2021open, minderer2023scaling, wang2022cris}. 

Inspired by the success of VLMs in natural image applications, researchers have adapted these models to the remote sensing domain. Recent efforts \cite{liu2024remoteclip, zhang2023rs5m, wang2024skyscript, pang2024h2rsvlm} have explored applying vision-language contrastive learning for aerial and satellite imagery, adapting general VLMs to remote sensing context with the constructed specialized cross-modal datasets. Therefore, these models encode domain-specific knowledge, enhancing their ability to extract and utilize remote sensing features effectively.
\section {Method}
\subsection{Task Formation}
Open-vocabulary remote sensing instance segmentation aims to segment an image $I\in \mathbb{R}^{H\times W\times 3}$ into a set of pixel-wise instance masks $m_i$ with associated aerial categories $c_i$:
\begin{align}
    \{y_i\}^K_{i=1} = \{(m_i, c_i)\}^K_{i=1}.
\end{align}
During training, the model learns from a predefined set of categories $\mathbf{C}_\mathrm{train}$, while during inference, it is evaluated on a separate set of categories $\mathbf{C}_\mathrm{test}$, where $\mathbf{C}_\mathrm{train} \neq \mathbf{C}_\mathrm{test}$. The category names of $\mathbf{C}_\mathrm{test}$ is available during testing.

\subsection{Framework Overview}
The overall framework of our proposed \ours is illustrated in Figure \ref{fig:framework}. The pipeline consists of three main parts: the context branch, semantic branch, and instance branch. In the context branch ({\color{pptblue}blue} arrows), we extract multi-granularity scene context with a remote-sensing (RS) CLIP model. The semantic branch ({\color{pptyellow}yellow} arrows) represents the classifier, which consists of text embeddings encoded by a frozen CLIP text encoder. The instance branch ({\color{pptorange}orange} arrows) is responsible for instance proposal generation. A frozen general CLIP image encoder extracts backbone features, which are then fed into the proposal generator to produce class embeddings and mask proposals. These three branches interact through the Region-Aware Integration (RAI) and Global Context Adaptation (GCA) modules. In RAI, regional context is integrated into class embedding,  enabling a richer understanding of surrounding environment. In GCA, domain-specific [CLS] token, \ie, global context, is injected into text embeddings, enhancing the classifier's adaptability for aerial objects. Finally the class prediction is derived from region-aware class embeddings $\hat{\mathbf{V}}$ and the domain adapted classifier $\hat{\mathbf{T}}$.

\begin{figure}[t]
  \centering
    \includegraphics[width=0.95\linewidth]{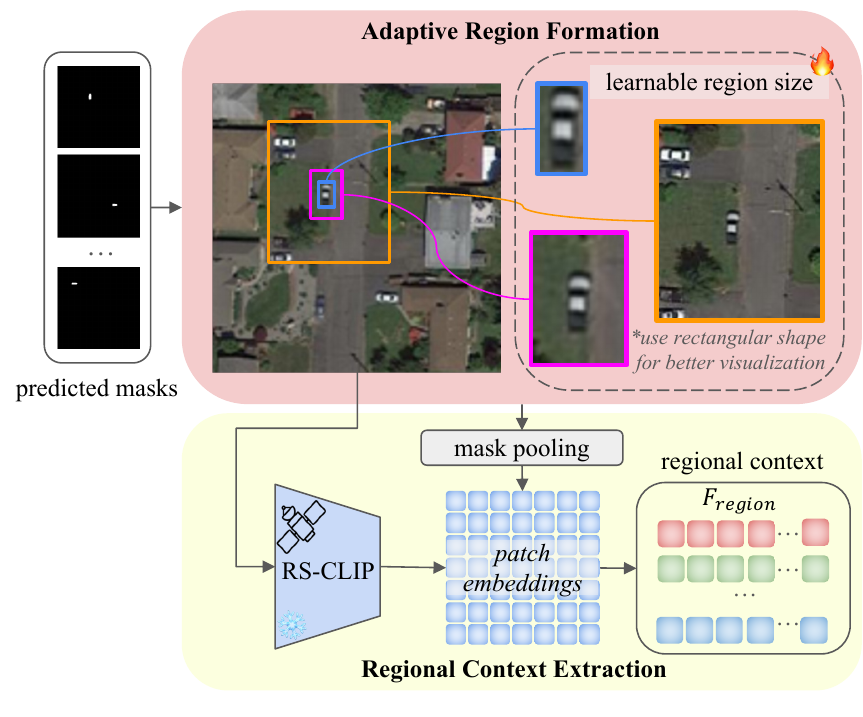}
   \caption{Illustration of Adaptive Region Formation and Regional Context Extraction in Region-Aware Integration.}
   \label{fig:sai}                                                                                                                                                                                                                                                                           
\end{figure}

\subsection{Scene Context Extraction}
To have a domain-specific knowledge about the scene, we utilize remote-sensing vision-language model, \ie, RemoteCLIP \cite{liu2024remoteclip}, to extract scene context from an input image $I\in \mathbb{R}^{H\times W\times 3}$. The ViT-based image encoder of RemoteCLIP, denoted as $\mathrm{CLIP}^*_\mathrm{RS}$ (* indicates frozen), extracts features as follow:
\begin{align}
    \{\mathbf{F}^i_\mathrm{CLS}, \mathbf{F}^i_{HW}\} = \mathrm{CLIP}^*_\mathrm{RS}(I),
\end{align}
where $i$ indicates the $i$-th transformer layer, $\mathbf{F}^i_\mathrm{CLS} \in \mathbb{R}^{1\times C}$ is the [CLS] token representing global image features, and $\mathbf{F}^i_{HW} \in \mathbb{R}^{\frac{H}{14}\times \frac{W}{14}\times C}$ contains patch embeddings encoding spatially dense image features. The final [CLS] token $\mathbf{F}^{final}_\mathrm{CLS}$ encapsulates the global context of the image, and the final patch embeddings $\mathbf{F}^{final}_{HW}$ provide rich contextual visual representations for the input image.

\subsection{Region-Aware Integration}
Due to the significant scale variations and resolution disparities in aerial images, it is challenging to accurately recognizing small or ambiguous aerial objects. However, remote sensing images contain valuable prior knowledge, as objects often correlate with their surrounding environments \cite{li2021robust}. To exploit this correlation, we propose Region-Aware Integration, which dynamically incorporates region-level scene context into class embeddings.

\vspace{-3mm}
\paragraph{Adaptive Region Formation.} Given $N$ class embeddings $\mathbf{V} = [v_1, v_2,... v_N] \in \mathbb{R}^{N\times C}$ and their corresponding predicted mask proposals $\mathbf{M} = [m_1, m_2,... m_N] \in \mathbb{R}^{B\times N\times H\times W}$, we suggest that the predicted masks $M$ serve as spatial references for each object instance. To get the surrounding region, we design an learnable dilation mechanism that adaptively expands predicted masks. We introduce a learnable dilation factor $\delta$ that adjusts the expansion dynamically. We define the expanded mask as:
\vspace{-6pt}
\begin{align}
    \mathbf{M}' = \max_{\mathbf{x}\in \mathcal{N}(M,k)}\mathbf{M}(\mathbf{x}),
\end{align}
where $\mathcal{N}(M,k)$ denotes the local neighborhood defined by a learnable kernel size $k$, computed as:
\begin{align}
    k = 3 + \mathrm{clamp}(\delta, 0, 10),
\end{align}
where $\delta$ is a learnable parameter initialized to 1 and optimized during training. The expansion is performed via max-pooling with kernel size $k$, stride 1, and padding $k//2$ to ensure spatial alignment.

\vspace{-3mm}
\paragraph{Regional Context Extraction.}
Once the expanded masks are obtained, we leverage the patch embeddings $\mathbf{F}^{final}_{HW}$, extracted from RemoteCLIP, to encode scene-level semantics. These embeddings capture dense visual features across the input image. To extract the regional context for each query object, with the corresponding expanded mask $m_i'$, we pool all the features in $\mathbf{F}^{final}_{HW}$ that are within the expanded mask to compute a mask pooled image feature as follows:
\begin{align}
    F_{\mathrm{region}} = \sum_{\mathbf{x}\in \mathbf{M'}}\omega_{\mathbf{x}} \cdot \mathbf{F}^{final}_{HW},
\end{align}
with ${\omega_\mathbf{x}}$ denotes the normalized weight within the mask $m_i'$. The adaptive region formation and context extraction process is illustrated in Figure \ref{fig:sai}.

\vspace{-3mm}
\paragraph{Regional Context Integration.}
With the extracted regional context $F_{region}$, we then integrate it into the class embeddings $\mathbf{V}$ through $l$ sequential Transformer Layers:
\vspace{-6pt}
\begin{align}
    \mathbf{V}_{i+1} = \mathrm{TransLayer}_i(\mathbf{V}_i, \lambda \cdot \mathbf{F}_\mathrm{region}),
\end{align}
where $i=1, 2, ... l$ and $\lambda$ is a temperature coefficient that controls the contribution of the regional context. Finally, we get region-aware class embeddings $\hat{\mathbf{V}}$.

This process seamlessly inject the regional scene context into class embeddings, utilizing domain-specific knowledge pretrained in remote sensing-specific CLIP. By leveraging the correlation between objects and their surrounding environments, the model gains a richer contextual understanding, improving its ability to distinguish aerial objects. Additionally, integrating scene information helps recalibrate the feature space of class embeddings trained on seen categories $\mathbf{C}_\mathrm{train}$, aligning them with a more generalized latent space. This adjustment mitigates overfitting to training categories, enabling better generalization to novel, unseen objects in open-vocabulary remote sensing instance segmentation.

\subsection{Global Context Adaptation}
Associating predicted instances with their regional context improves their visual distinguishability, yet the performance is still limited by the frozen classifier built on CLIP text embeddings. Although the classifier, derived from general-domain CLIP, exhibits strong generalization abilities, its fixed text embeddings lack the specificity needed to accurately distinguish remote sensing objects. This limitation arises from factors such as high intra-class variation, resolution differences, and complex environmental conditions. To address this, we complement Region-Aware Integration, which enhances the visual modality, by further refining the textual modality. Our approach bridges the semantic gap between general-domain knowledge and remote sensing-specific concepts, which improves the classifier's adaptability for open-vocabulary remote sensing instance segmentation. Specifically, we incorporate global scene context to adapt domain-specific visual priors into the text embeddings, ensuring a more robust alignment between visual and textual representations.

The classifier $\mathbf{T}$ consists of text embeddings encoded by the frozen CLIP text encoder. Given a set of category names $\mathbf{C}_\mathrm{train}$ represented in natural language, we derive category text embeddings by place the category name into prompt templates, \eg, ‘satellite imagery of {}.’, ‘aerial imagery of {}.’, and then fed into CLIP text encoder. The text embedding for each class is the average across all templates. We denote the classifier as $\mathbf{T}\in \mathbb{R}^{M\times C}$, where $M$ is the number of categories, $C$ is the dimension of text embeddings.

To equip the classifier with remote sensing knowledge, global visual context, \ie, $\mathbf{F}^{final}_\mathrm{CLS}$ from RemoteCLIP is injected into the text embeddings using multi-head cross-attention with $\mathbf{T}$, formulated as:
\begin{equation}
    \begin{aligned}
    Q=\mathbf{w}_Q\mathbf{F}^{final}_\mathrm{CLS},K=\mathbf{w}_K\mathbf{T}, V=\mathbf{w}_V\mathbf{T},\\
    \hat{\mathbf{T}} =\mathrm{MHA}(Q, K, V) = \mathrm{softmax}(\frac{QK^T}{\sqrt{d_k}})V,
    \end{aligned}
\end{equation}
where $\mathbf{w}_Q, \mathbf{w}_K, \mathbf{w}_V$ are linear projection matrices that map inputs to the shared feature dimension $d_k$. This adaptation process ensures that the text embeddings retain their open-vocabulary generalization while being enriched with remote sensing-specific global context.

By integrating domain-specific global context into the text embeddings, our approach facilitates cross-domain interactions to produce more representative and adaptable textual features tailored for aerial images. Since these context are extracted from visual representations, the method also strengthen cross-modal interactions between vision and language, promoting better semantic alignment. As a result, the classifier becomes more robust in handling the variability of remote sensing objects, ultimately leading to improved performance across diverse datasets.

\subsection{Open-Vocabulary Inference}
During inference, the classification score is computed using an ensemble approach \cite{xu2023open, yu2023convolutions} that combines both in-vocabulary and out-vocabulary classification. The in-vocabulary classification is based on the learned region-aware class embeddings and then classified with text embeddings enhanced with domain global context. For out-vocabulary classification, our framework offers flexibility in implementation. By leveraging the instance branch (built on a general CLIP backbone) and the context branch (using a remote sensing-specific CLIP backbone), dense visual features can be extracted from both general and remote sensing domains. While the use of remote sensing CLIP models appears promising, our experimental results in Section \ref{sec:exp_abl} indicate that they still fall short compared to general-domain CLIP in our task. This observation aligns with the results reported in \cite{ye2025GSNet}. As a result, we continue to utilize the general CLIP to get out-vocabulary classification due to its superior generalization capabilities.

\section{Experiments}
\label{sec:exp}
\subsection{Datasets and Evaluation Metric}

\paragraph{Training Dataset.}
Following the open-vocabulary benchmarks for natural images \cite{xu2022simple}, we train the model on one dataset and evaluate its cross-dataset performance on other datasets. We select two datasets for training, \ie, iSAID \cite{zamir2019isaid} and SIOR \cite{wang2023samrs}. iSAID is a large scale instance segmentation dataset for remote sensing images. It contains 18732 images for training across 15 categories. SIOR is developed from aerial object detection dataset DIOR \cite{li2020object}, with segmentation annotations generated in SAMRS \cite{wang2023samrs}. It contains 11725 images with 20 aerial categories. 

\setlength{\tabcolsep}{8pt}
\begin{table*}[t]
\footnotesize
  \begin{center}
  \begin{tabular}{lcccccccccc}
  \toprule
    & \multicolumn{5}{c}{iSAID} & \multicolumn{5}{c}{SIOR}\\
  \cmidrule(r){2-6} \cmidrule(r){7-11}
   Method & NWPU  & SOTA & FAST & SIOR & Average & NWPU  & SOTA & FAST & iSAID & Average\\
  \cmidrule(r){2-6} \cmidrule(r){7-11}
 ODISE \cite{xu2023open} \textcolor{gray}{[CVPR22]}& 36.40 & 13.91 & 4.65 & 13.68 &	17.16 & 41.77 &	12.3 & 5.88 & 12.68 & 18.16\\
FC-CLIP \cite{yu2023convolutions} \textcolor{gray}{[NeurIPS23]} & 60.67 & 33.62 &	11.88 &	26.79 & 33.24 & 60.69 & 19.84 &	8.67 &	22.24 & 27.86\\
MAFT+ \cite{jiao2024collaborative} \textcolor{gray}{[ECCV24]}&  35.32 &	6.63 & 5.52 &	9.84 &	14.33 & 39.99  & 7.77 &	5.92 &	9.63 & 15.83 \\
ZoRI \cite{huang2024zori} \textcolor{gray}{[AAAI25]}&  62.06 & 30.02 & 12.65 &	26.27 &	32.75 & 59.77 &	20.26 &	9.58 & 23.46 & 28.27 \\
\rowcolor{gray!20} \ours \textcolor{gray}{(Ours)} & \textbf{67.59}  &  \textbf{42.57} &  \textbf{13.67}  &  \textbf{30.90}  &  \textbf{38.68} &  \textbf{69.17}  &  \textbf{23.68}  &  \textbf{10.33} &   \textbf{27.15} &   \textbf{32.59}  \\
    \toprule
  \end{tabular}
\caption{\textbf{Comparison with SOTA methods on open-vocabulary remote sensing instance segmentation.} The model is trained separately on iSAID and SIOR datasets and then tested on the remaining four datasets to measure its cross-dataset generalization capabilities.}
\label{tab:sota}
\end{center}
\vspace{-6pt}
\end{table*}

\setlength{\tabcolsep}{10pt}
\begin{table*}[t]
\footnotesize
  \begin{center}
  \begin{tabular}{cccccccccccc}
  \toprule
   \multicolumn{2}{c}{Modules} & \multicolumn{5}{c}{iSAID} & \multicolumn{5}{c}{SIOR}\\
  \cmidrule(r){1-2} \cmidrule(r){3-7} \cmidrule(r){8-12}
  RAI & GCA & NWPU  & SOTA & FAST & SIOR & Average & NWPU  & SOTA & FAST & iSAID & Average\\
  \cmidrule(r){1-2} \cmidrule(r){3-7} \cmidrule(r){8-12}
\xmarkg& \xmarkg &  58.59  &  36.44  &  11.56  &  26.43  &  33.25 &  61.02  &  20.85  &  9.14  &  22.44  &  28.36\\
\cmark& \xmarkg & 66.32  &  39.55 & 12.85 &  28.91   &  36.91  &  68.47  & 21.28  &  9.70 &  25.16  &  31.15\\
\xmarkg& \cmark &  67.21  &  38.14 &  12.37  &  28.96   &  36.67 &  69.07 &  22.05 & 9.79 &  24.61 &  31.43 \\
\rowcolor{gray!20} \cmark& \cmark&  \textbf{67.59}  &  \textbf{42.57} &  \textbf{13.67}  &  \textbf{30.90}  &  \textbf{38.68} &  \textbf{69.17}  &  \textbf{23.68}  &  \textbf{10.33} &   \textbf{27.15} &   \textbf{32.59} \\
    \toprule
  \end{tabular}
\caption{\textbf{Component analysis of \ours.} The cross-dataset evaluation results based on models trained on the iSAID and SIOR datasets are provided. RAI denotes Region-Aware Integration and GCA denotes Global Context Adaptation.}
\label{tab:component_analysis}
\vspace{-12pt}
\end{center}
\end{table*}

\vspace{-3mm}
\paragraph{Evaluation Dataset.}
To evaluate the effectiveness of our method, we conduct cross-dataset evaluation on 4 aerial instance segmentation datasets, \ie, NWPU-VHR-10 \cite{cheng2014multi,su2019object}, SOTA \cite{wang2023samrs}, FAST \cite{wang2023samrs}, and SIOR \cite{wang2023samrs}. NWPU-VHR-10 is an aerial object detection dataset with instance masks further annotated by \cite{su2019object}. The test set contains 731 images from 10 aerial classes. SOTA, FAST and SIOR are segmentation datasets provided in SAMRS \cite{wang2023samrs}, which are developed from aerial object detection datasets DOTA-V2.0 \cite{ding2021object}, FAIR1M-2.0 \cite{sun2022fair1m}, and DIOR \cite{li2020object}, respectively. SOTA covers 874 images with 18 object categories for testing, FAST contains 3207 images across 37 fine-grained aerial categories for testing, and SIOR comprises 11738 testing samples across 20 classes. Additional dataset details can be found in the supplementary material.

\vspace{-3mm}
\paragraph{Evaluation Metric.}
We follow previous work \cite{qin2023freeseg} to evaluate the performance with mean Average Precision (mAP) for open vocabulary instance segmentation.

\subsection{Implementation Details}
For instance branch, we use frozen ConvNeXt-Large CLIP as the backbone \cite{liu2022convnet2020s} with the weight pretrained on LAION-2B from OpenCLIP \cite{ilharco_gabriel_2021_5143773}. Its generalization to high-resolution inputs is better suited to extract features for segmentation instead of using a ViT-based CLIP. The proposal generator follows Mask2Former \cite{cheng2022masked} with number of object query set to 300. In semantic branch, prompt templates for remote sensing images RESISC45 \cite{cheng2017remote} 
are employed to obtain text embeddings with the pretrained CLIP text encoder. For context branch, we employ RemoteCLIP ViT-L/14 \cite{liu2024remoteclip} to extract multi-granularity scene context. All ablation experiments are trained for 50 epochs with batch size set to 2, on one L40S GPU. The learning rate is set to $1.25\times 10^{-5}$.
Input images are resized to $512\times 512$ during training. The model is optimized using AdamW optimizer.

\subsection{Comparison with State-of-the-Art Methods}
Table \ref{tab:sota} presents a comparative analysis of our method \ours against SOTA models training on the iSAID and SIOR datasets respectively. Our approach consistently outperforms existing methods across all benchmarks, demonstrating its effectiveness in open-vocabulary remote sensing instance segmentation. Specifically, we achieve an average  improvement of 5.53\% and 4.32\% over the best-performing SOTA models
trained on iSAID and SIOR, respectively. Our method surpasses previous methods by a large margin, especially on NWPU, SOTA, SIOR (trained on iSAID) and iSAID (trained on SIOR) datasets, with gains up to 12.55\%.

\subsection{Ablation Studies}
\label{sec:exp_abl}
We perform a series of ablation experiments to assess the effectiveness of our proposed modules. The component analysis presents cross-dataset evaluation results based on models trained on the iSAID and SIOR datasets. Meanwhile, the remaining ablation studies are conducted solely on experiments where iSAID serves as the training dataset.

\vspace{-4mm}
\paragraph{Component Analysis.}
The results, as shown in Table \ref{tab:component_analysis}, highlight the contributions of Region-Aware Integration (RAI) and Global Context Adaptation (GCA) modules, which are central to facilitating OV remote sensing instance segmentation.
When RAI is enabled, the mAP increases nearly 8\% for NWPU dataset trained in iSAID and SIOR respectively, compared to the baseline configuration. We can also observe an average performance gain of around 3 \% for the remaining three datasets. This improvement underscores the effectiveness of incorporating regional context to refine class embeddings, retrieving region-aware class embedding and calibrating the feature to a more generalizable space. Additionally, VCA achieves an average improvement of 3.42\% on iSAID and 3.07\% on SIOR. These results validate our approach of enriching the linguistic latent space with global visual context, making the classifier more adaptable to remote sensing objects. By combining both RAI and VCA, our models yields the best performance, demonstrating their complementary roles: RAI refines class embedding with regional context in the visual space, while VPA enhances text embeddings by incorporating global scene context.

\vspace{-4mm}
\paragraph{VLMs for Scene Context Extraction.}
As mentioned in the related work \ref{sec:related_work_vlm}, several remote sensing (RS) CLIPs have been developed recently. We evaluate the impact of different VLMs for scene context extraction, comparing the general CLIP \cite{radford2021learning} with three popular remote sensing-specific CLIPs, \ie, SkyCLIP \cite{wang2024skyscript}, GeoRSCLIP \cite{zhang2023rs5m}, and RemoteCLIP \cite{liu2024remoteclip}. To ensure a fair comparison, we use the ViT-L/14 pretrained weights across all models. As shown in Table \ref{tab:vlm}, RS CLIPs outperform the general CLIP on almost all test datasets. This highlights the advantages of domain-specific pretraining in capturing more accurate and specialized scene context for aerial images. Among them, RemoteCLIP achieves the best performance and is therefore employed in our framework.
\setlength{\tabcolsep}{9pt}
\begin{table}[t]
\footnotesize
  \begin{center}
  \begin{tabular}{l|cccc}
  \shline
VLM  & NWPU  & SOTA & FAST & SIOR \\
  \shline
CLIP \cite{radford2021learning} &  64.03 & 38.74 & 11.42	& 28.11  \\
SkyCLIP \cite{wang2024skyscript}& 65.04 & 33.57	& 12.43	& 28.96  \\
GeoRSCLIP \cite{zhang2023rs5m}& 64.72	& 39.33 & 12.56	& 28.13 \\
\rowcolor{gray!20}RemoteCLIP \cite{liu2024remoteclip} & \textbf{67.59} &  \textbf{42.57} &  \textbf{13.67}  &  \textbf{30.90}\\
\shline
  \end{tabular}
\caption{\textbf{Ablation Study -- VLMs for Scene Context Extraction.} Comparison of various VLMs used in the context branch for multi-granularity scene context extraction. We include general CLIP and three RS CLIPs for comparison.}
\vspace{-4mm}
\label{tab:vlm}
\end{center}
\end{table}

\vspace{-4mm}
\paragraph{Choice of Context in RAI.}
\setlength{\tabcolsep}{10pt}
\begin{table}[t]
\footnotesize
  \begin{center}
  \begin{tabular}{l|cccc}
  \shline
 Choice of Context & NWPU  & SOTA & FAST & SIOR \\
  \shline
$\mathrm{[CLS]}$ token & 67.06  &  40.35 &  12.58  &  29.07  \\
Patch embeddings & 66.14 &  37.95  & 12.60  &  28.97 \\
\rowcolor{gray!20} Regional context &  \textbf{67.59}  &  \textbf{42.57} &  \textbf{13.67}  &  \textbf{30.90}  \\
\shline
  \end{tabular}
\caption{\textbf{Ablation Study -- Choice of Context in RAI}. Comparison of different context that can be used in Region-Aware Integration.}
\vspace{-20pt}
\label{tab:sai}
\end{center}
\end{table}

We explore various context that can be applied for RAI module in Table \ref{tab:sai}. As a starting point, we utilize the [CLS] token from RemoteCLIP, which encapsulates the global context of the input image. The performance gain over the baseline supports our intuition that integrating scene context into class prediction can help identify instances in remote sensing imagery. However, it still falls short compared to our proposed regional context. This could be attributed to the diverse concepts and semantics that can include in large-scale remote sensing imagery. Since the [CLS] token extracts high-level semantic features, it tends to focus on the dominant components within the image, introducing a global bias \cite{li2024segearth}. As a result, the model may struggle to accurately predict less prominent objects. Additionally, intermediate patch embeddings $\mathbf{F}^i_{HW}$ yield the lowest performance, likely because mid-layer representations emphasize texture over semantic information, introducing unrelated noise to the process. By contrast, our regional context, which is adaptively formulated, allows the model to leverage the local context of the target object while mitigating the effects of global bias and noise.

\begin{figure*}[t]
  \centering
    \includegraphics[width=0.95\textwidth]{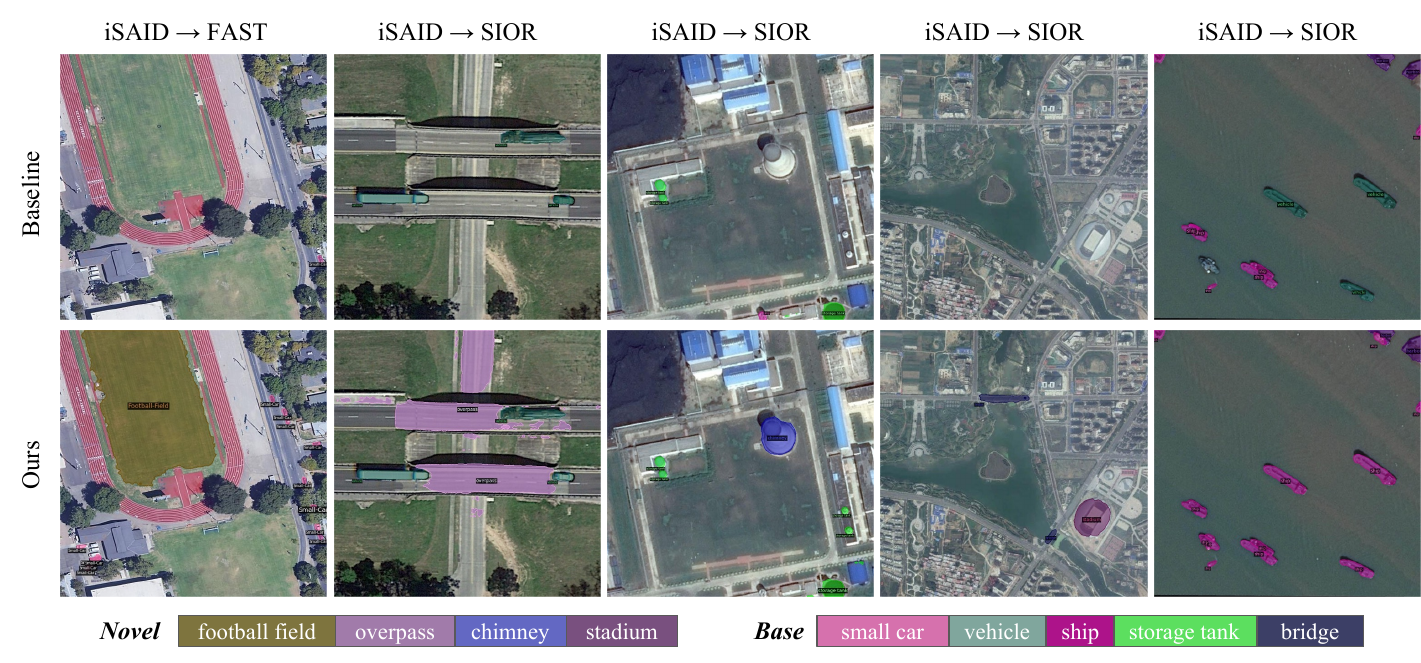}
    \vspace{-3pt}
   \caption{\textbf{Qualitative comparisons between the baseline and our model.} ``iSAID $\rightarrow$ FAST'' denotes training on iSAID and testing on FAST. We set $\texttt{jittering=False}$ for better readability. \textbf{\textit{Novel}} and \textbf{\textit{Base}} indicate whether a class is absent or present in the training dataset, respectively. \ours effectively segments novel classes such as \textit{football field} (column 1), \textit{overpass} (column 2), \textit{chimney} (column 3), \textit{stadium} (column 4). Furthermore, our model demonstrates strong instance segmentation capabilities in large-scale images, successfully segmenting \textit{small car} (column 1) and \textit{stadium} (column 4). Additionally, in the last column, we correctly identify all \textit{ship} instances, whereas the baseline model confuses them with \textit{vehicle}. }
   \label{fig:pic}
   \vspace{-9pt}
\end{figure*}

\vspace{-4mm}
\paragraph{Adaptation Methods in GCA.}
\setlength{\tabcolsep}{11pt}
\begin{table}[t]
\footnotesize
  \begin{center}
  \begin{tabular}{l|cccc}
  \shline
 Method & NWPU  & SOTA & FAST & SIOR\\
  \shline
w/o injection &  66.32 & 39.55 & 12.85 & 28.91 \\
add &  65.72  & 39.89  & 11.11  &  29.86  \\
concat & 47.53 &  15.51  & 1.74 & 18.23  \\
\rowcolor{gray!20} MHA &  \textbf{67.59}  &  \textbf{42.57} &  \textbf{13.67}  &  \textbf{30.90} \\
\shline
  \end{tabular}
\caption{\textbf{Ablation Study -- Adaptation Methods in GCA.} Comparison of different global visual context injection methods in Global Context Adaptation.}
\label{tab:vca}
\end{center}
\end{table}
To inject remote sensing-specific global context into the text embedding, different integration methods can be used. As listed in Table \ref{tab:vca}, we experiment with straightforward approaches such as addition and concatenation, both of which do not involve trainable parameters. The results indicate that these methods fail to produce a generalizable classifier, even performing worse than the naive classifier without global context injection. We attribute this to the misalignment between the RS CLIP visual embeddings and the general CLIP textual embeddings.  Simply combining these two without learnable parameters may disrupt the well-pretrained cross-modal alignment. Instead, by treating the global context as a query and allowing it to cross-attend to the text embeddings $\mathbf{T}$, we facilitate a smoother interaction between the specialized and general CLIP representations.

\vspace{-4mm}
\paragraph{Out-Vocabulary Classification.}
\setlength{\tabcolsep}{4pt}
\begin{table}[t]
\footnotesize
  \begin{center}
  \begin{tabular}{l|c|cccc}
  \shline
Context VLM & OV VLM & NWPU & SOTA & FAST & SIOR  \\
  \shline
   \multirow{2}{*}{SkyCLIP \cite{wang2024skyscript}} 
   & SkyCLIP & 64.57	& 32.49  &	12.44	& 27.22 \\
   \cline{2-6}
   &\cellcolor{gray!20}  CLIP & \cellcolor{gray!20} 65.04 & \cellcolor{gray!20} 33.57 & \cellcolor{gray!20} 12.42 & \cellcolor{gray!20} 28.96 \\
   \shline
   \multirow{2}{*}{GeoRSCLIP \cite{zhang2023rs5m}} 
   & GeoRSCLIP & 63.67 & 38.76 & 12.55 & 27.15 \\
   \cline{2-6}
   & \cellcolor{gray!20} CLIP & \cellcolor{gray!20} 64.72 & \cellcolor{gray!20} 39.33 & \cellcolor{gray!20} 12.56 & \cellcolor{gray!20} 28.13 \\
   \shline
   \multirow{2}{*}{RemoteCLIP \cite{liu2024remoteclip}} 
   & RemoteCLIP & 66.05 &	40.26 & 11.65 &	29.43 \\
   \cline{2-6}
   & \cellcolor{gray!20} CLIP & \cellcolor{gray!20} 67.59 &  \cellcolor{gray!20} 42.57 & \cellcolor{gray!20} 13.67  &  \cellcolor{gray!20} 30.90 \\
    \shline
  \end{tabular}
\caption{\textbf{Ablation Study -- Open-Vocabulary Classification.} The context VLM denotes different RS CLIPs used for scene context extraction in the context branch, while OV VLM refers to the model used for open-vocabulary classification. The results mainly investigate the impact of employing domain-specific CLIP versus general CLIP in open-vocabulary classification.}
\vspace{-20pt}
\label{tab:classifier}
\end{center}
\end{table}
During inference, out-vocabulary (OV) classification is used to further enhance novel class prediction. Our framework employs RS CLIP for scene context extraction in the context branch and general CLIP as the backbone in the instance branch. Thus, we provide flexible options for OV classification. As shown in Table \ref{tab:classifier}, the context VLM refers to different RS CLIPs used for scene context extraction, while OV VLM represents the models from which our OV classification is derived.  We conduct three sets of experiments based on different RS CLIPs used for scene context extraction. The trained model is then evaluated with (1) the same context VLM as the OV VLM (2) general CLIP as the OV VLM. From the results, we can see that, although RS CLIPs can produce reasonable performance, they consistently underperform general-domain CLIP in our task. This observation aligns with the results provided in \cite{ye2025GSNet} for OV semantic segmentation. Consequently, we adopt general CLIP to get OV classification probability, which is then combine with in-vocabulary classification to yield the final prediction. We think the performance gap primarily comes from the limited availability of image-caption pairs for cross-modal pretraining in RS CLIPs. The remote sensing pretraining datasets in these models range from 0.8M to 5M image-text pairs, which is considerable smaller than 400M web-scale data used in general CLIP. As a result, RS CLIPs acquire domain-specific knowledge but exhibit weaker generalization capabilities.

\vspace{-4mm}
\paragraph{Qualitative Results.}
In Figure \ref{fig:pic}, we provide visualizations of our performance on remote sensing instance segmentation task. \ours effectively segments the instances of novel classes such as \textit{football field} (column 1), \textit{overpass} (column 2), \textit{chimney} (column 3) and \textit{stadium} (column 4) across different test datasets, demonstrating strong class prediction capabilities through the synergy of RAI and VCA modules. Furthermore, our model exhibits robustness in segmenting small instances in large-scale images. For instance, in the 1st column, most \textit{small car} instances are segmented, whereas the baseline fails to segment any. In the 4th column, we successfully segment both novel class \textit{stadium} and base class \textit{bridge}, even in an extremely large-scale image captured from a high altitude. Additionally, in the last column, many instances of \textit{ship} are misclassified as \textit{vehicle} by the baseline model, whereas \ours correctly identifies all of them as \textit{ship}. This aligns with our motivation that surrounding context aids in aerial object recognition. More cross-dataset qualitative results are provided in the supplementary material.

\section{Conclusion}
\vspace{-1mm}
In conclusion, we  introduce the task of open-vocabulary remote sensing instance segmentation and develop \ours, a comprehensive framework that integrates domain-specific scene context. By leveraging the correlation between objects and their surroundings, RAI incorporates regional context to class embeddings to gain richer contextual understanding. GCA injects domain-specific global context into text embeddings to create a more adaptive classifier for remote sensing objects. Extensive experiments validate the effectiveness of our approach, and with its state-of-the-art performance, we provide a reliable solution for large-scale real-world geospatial analysis.

\paragraph{Acknowledgments.}
This research is supported in part by the National Research Foundation Singapore Competitive Research Program (award number CRP29-2022-0003). The work was done at Rapid-Rich Object Search (ROSE) Lab, School of Electrical $\&$ Electronic Engineering, Nanyang Technological University. Shuting He is sponsored by Shanghai Pujiang Programme 24PJD030 and Natural Science Foundation of Shanghai 25ZR1402138.  


{
    \small
    \bibliographystyle{ieeenat_fullname}
    \bibliography{main}
}

\clearpage
\setcounter{page}{1}
\maketitlesupplementary

\renewcommand{\thesection}{\Alph{section}}
\renewcommand{\thetable}{\Alph{table}}
\renewcommand{\thefigure}{\Alph{figure}}
\renewcommand{\theequation}{\Alph{equation}}

\setcounter{section}{0}
\setcounter{table}{0}
\setcounter{figure}{0}
\setcounter{equation}{0}

In the supplementary materials, we provide more information on the datasets used for open-vocabulary remote sensing instance segmentation benchmark and include more qualitative results along with comparisons. Moreover, we show that \ours can also enhance the performance for open-vocabulary remote sensing semantic segmentation task, which further unleashes the potential of our model.

\section{Implementation Details}
\subsection{Remote Sensing Instance Segmentation}

\paragraph{Training Dataset.}
Following the open-vocabulary benchmarks for natural images \cite{xu2022simple}, we train the model on one dataset and evaluate its cross-dataset performance on other datasets. 
We select two datasets for training, \ie, iSAID \cite{zamir2019isaid} and SIOR \cite{wang2023samrs}.
iSAID is a large scale instance segmentation dataset for remote sensing images. 
It contains 18732 images for training across 15 categories. 
SIOR is developed from aerial object detection dataset DIOR  \cite{li2020object}, with segmentation annotations generated in SAMRS \cite{wang2023samrs}, which contains 11725 images with 20 categories.  

\paragraph{Evaluation Dataset.}
To evaluate the effectiveness of our
method, we conduct cross-dataset evaluation on 4 aerial instance segmentation datasets, \ie, NWPU-VHR-10 \cite{cheng2014multi,su2019object}, SOTA \cite{wang2023samrs}, FAST \cite{wang2023samrs}, and SIOR \cite{wang2023samrs}. 
NWPU-VHR-10 is an aerial object detection dataset with instance masks further annotated by \cite{su2019object}. 
The test set contains 731 images from 10 aerial classes. SOTA, FAST and SIOR are segmentation datasets provided in SAMRS \cite{wang2023samrs}, which are developed from aerial object detection datasets DOTA-V2.0 \cite{ding2021object}, FAIR1M-2.0 \cite{sun2022fair1m}, and DIOR \cite{li2020object}, respectively. 
SOTA covers 874 images with 18 object categories for testing, FAST contains 3207 images across 37 fine-grained aerial object categories for testing, and SIOR is with 11738 testing samples.
We provide the categories in each dataset in~\Cref{tab:dataset-category}.

\subsection{Remote Sensing Semantic Segmentation}

\paragraph{Training Dataset.}
Following the open-vocabulary benchmarks for remote sensing semantic segmentation~\cite{ye2024towards}, we train the model on their proposed LandDiscover50K dataset.
It includes 51846 high-resolution remote sensing images annotated across 40
object categories.

\paragraph{Evaluation Dataset.}
To evaluate the effectiveness of our method, we follow the evaluation settings in~\cite{ye2024towards} to conduct cross-dataset evaluation on 4 remote sensing semantic datasets, \ie, FLAIR~\cite{garioud2023flair}, FAST~\cite{wang2023samrs}, Potsdam~\cite{ISPRS_Potsdam}, and FloodNet~\cite{floodnet}.
Each dataset has its own bias towards different remote sensing categories.
To illustrate, Potsdam~\cite{ISPRS_Potsdam} emphasizes the in-vocabulary performance with high category similarity to the training LandDiscover50K dataset, which contains 5472 images with 6 semantic categories.
FloodNet~\cite{floodnet} focuses more on the post-flood analysis, which contains 898 images with 9 semantic categories.
FLAIR~\cite{garioud2023flair} is with 15700 images focusing on 12 large-scale landcover types.
FAST~\cite{wang2023samrs} contains 3207 images, specializing in 37 fine-grained semantic classes for remote sensing.
The combination of the four datasets enables a comprehensive evaluation of the open-vocabulary semantic segmentation tasks in remote sensing.
We provide the categories in each dataset in~\Cref{tab:dataset-category-semseg}.

\section{Additional Experiment Results}
\subsection{Additional Results on Semantic Segmentation}
The proposed \textbf{SCORE} can also be applied to diverse segmentation related tasks, \eg semantic segmentation.
We provide the semantic segmentation results of our method in~\Cref{tab:sota-semseg}.
Our approach consistently outperforms existing across three of four benchmarks, demonstrating its effectiveness in open-vocabulary remote sensing semantic segmentation.
Specifically, we achieve an average improvement of 1.13\% over the current SOTA model~\cite{ye2025GSNet}.
Our method surpasses previous methods by a large margin, especially on FLAIR and FAST datasets, with gains up to 9.62\%.

\setlength{\tabcolsep}{3pt}
\begin{table}[htbp]
\footnotesize
  \begin{center}
  \begin{tabular}{lccccc}
  \toprule
    & \multicolumn{5}{c}{LandDiscover50K }\\
  \cmidrule(r){2-6}
   Method & FLAIR  & FAST & Potsdam & FloodNet  & Average\\
  \cmidrule(r){2-6}
CAT-SEG \cite{cho2024cat} \textcolor{gray}{[CVPR24]}& 19.71 & 15.55 & 39.57 & 35.91 &	27.69 \\
GSNet \cite{ye2025GSNet} \textcolor{gray}{[AAAI25]}&  18.35 & 15.21 & \textbf{43.29} & 37.68 & 28.63 \\
\rowcolor{gray!20} \ours \textcolor{gray}{(Ours)} & \textbf{29.33}  &  \textbf{21.51} &  26.51  &  \textbf{41.70}  &  \textbf{29.76}  \\
    \toprule
  \end{tabular}
\caption{\textbf{Comparison with SOTA methods on open-vocabulary remote sensing semantic segmentation.} The model is trained on LandDiscover50K dataset and then tested on the four evaluation benchmarks to measure its cross-dataset generalization capabilities.}
\label{tab:sota-semseg}
\vspace{-9pt}
\end{center}
\end{table}

\subsection{Additional Qualitative Results}
\begin{figure*}[t]
  \centering
    \includegraphics[width=0.8\textwidth]{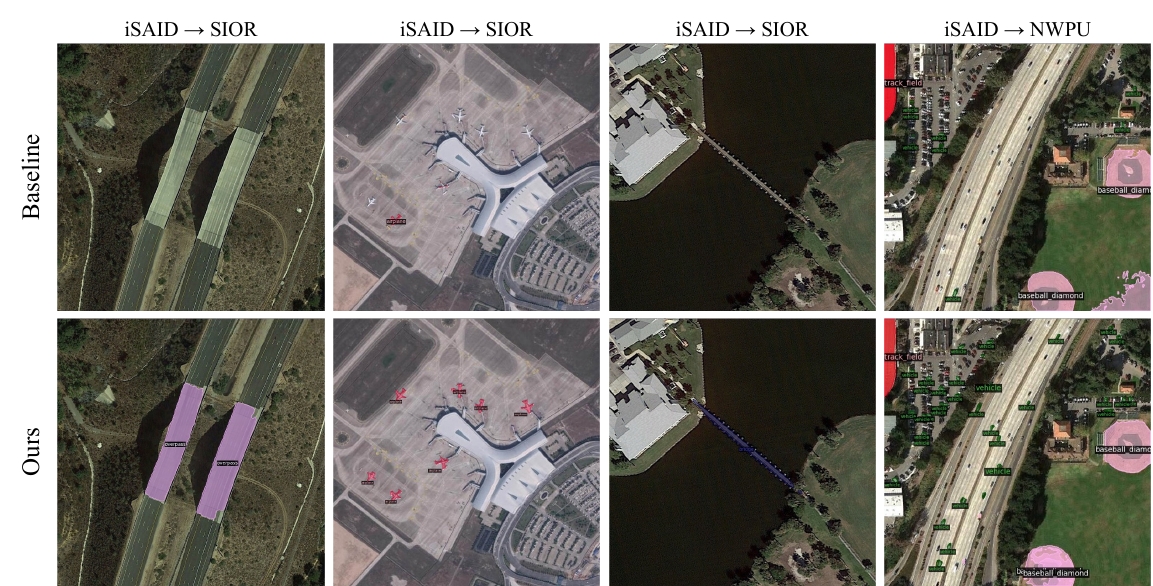}
    \includegraphics[width=0.8\textwidth]{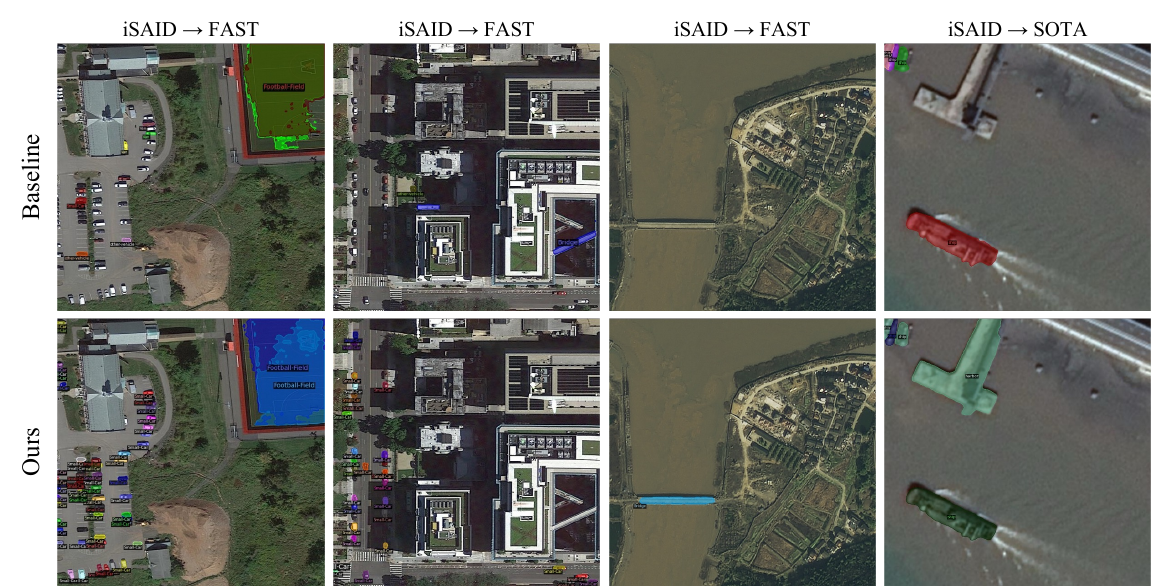}
    \includegraphics[width=0.8\textwidth]{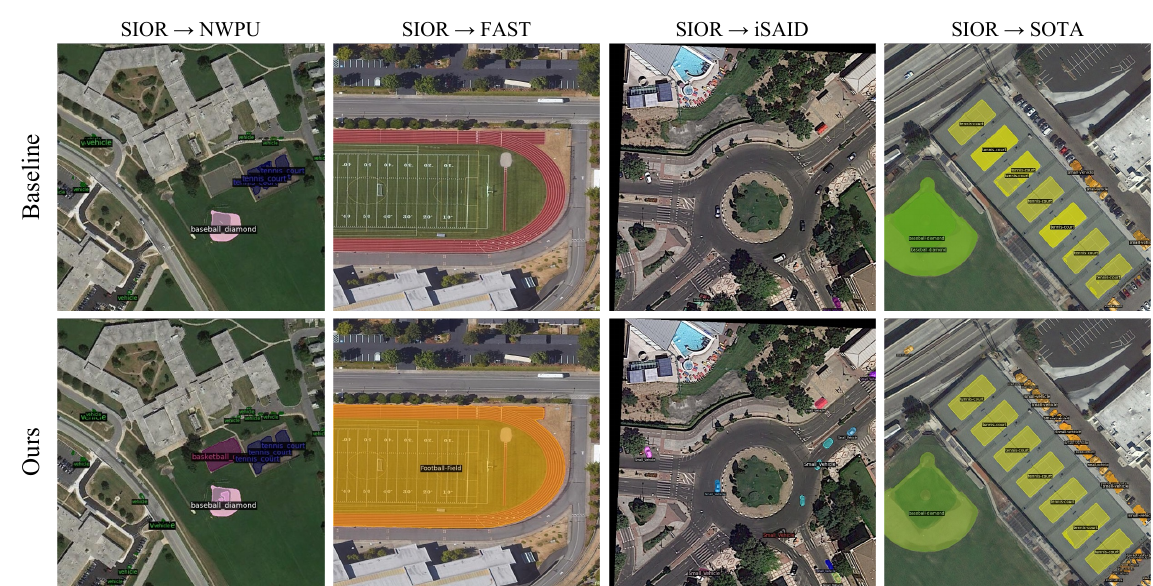}
   \caption{\textbf{Additional qualitative results between the baseline and our model.}}
   \label{fig:pic-more-qual}
   \vspace{-3pt}
\end{figure*}

We provide additional qualitative results of our proposed method on remote sensing instance segmentation task as shown in~\Cref{fig:pic-more-qual}.

\begin{table*}[t]

\centering
 \begin{adjustbox}{max width=\textwidth}
\begin{tabular}{lcl}
\toprule
Dataset             & \#Category & Category Name    \\
\hline
iSAID~\cite{zamir2019isaid} & 15 & \begin{tabular}[c]{@{}l@{}} ship, storage tank, baseball diamond, tennis court, basketball court, \\ground track field, bridge, large vehicle, small vehicle, \\helicopter, swimming pool, roundabout, soccer ball field, plane, harbor \end{tabular} \\  \hline
SIOR~\cite{wang2023samrs} & 20 & \begin{tabular}[c]{@{}l@{}} airplane, airport, baseball field, basketball court, bridge, chimney, \\expressway service area, expressway toll station, dam, golffield, \\ground track field, harbor, overpass, ship, stadium, \\storage tank, tennis court, train station, vehicle, windmill \end{tabular}\\  \hline
NWPU~\cite{cheng2014multi,su2019object} & 10 & \begin{tabular}[c]{@{}l@{}} airplane, ship, storage tank, baseball diamond, \\tennis court, basketball court, ground track field, harbor, bridge, vehicle\end{tabular}\\  \hline
FAST~\cite{wang2023samrs} & 37 & \begin{tabular}[c]{@{}l@{}} A220, A321, A330, A350, ARJ21, baseball field, basketball court, \\Boeing737, Boeing747, Boeing777, Boeing787, bridge, bus, C919, cargo truck, \\dry cargo ship, dump truck, engineering ship, excavator, fishing boat, \\football field, intersection, liquid cargo ship, motorboat, other-airplane, \\other-ship, other-vehicle, passenger ship, roundabout, \\small car, tennis court, tractor, trailer, truck tractor, tugboat, van, warship\end{tabular}\\  \hline
SOTA~\cite{wang2023samrs} & 18 & \begin{tabular}[c]{@{}l@{}} large vehicle, swimming pool, helicopter, bridge, plane, ship, \\soccer ball field, basketball court, ground track field, small vehicle, baseball diamond, \\tennis court, roundabout, storage tank, harbor, container crane, airport, helipad\end{tabular}\\ 

\bottomrule
\end{tabular}
\end{adjustbox}
\vspace{0.1cm}
\caption{\textbf{Category Names for datasets used in our instance segmentation benchmarks.} }
\vspace{10mm}
\label{tab:dataset-category}
\end{table*}
\begin{table*}[t]

\centering
 \begin{adjustbox}{max width=\textwidth}
\begin{tabular}{lcl}
\toprule
Dataset             & \#Category & Category Name    \\
\hline
LandDiscover50K~\cite{ye2025GSNet} & 40 & \begin{tabular}[c]{@{}l@{}} background, bare land, grass, pavement, road, tree, water, \\agriculture land, buildings, forest land, barren land, urban land, \\large vehicle, swimming pool, helicopter, bridge, \\plane, ship, soccer ball field, basketball court, \\ground track field, small vehicle, baseball diamond, \\tennis court, roundabout, storage tank, harbor, \\container crane, airport, helipad, chimney, \\expressway service area, expresswalltoll station, dam, \\golf field, overpass, stadium, train station, vehicle, windmill \end{tabular} \\  \hline
FLAIR~\cite{garioud2023flair} & 12 & \begin{tabular}[c]{@{}l@{}} building, pervious surface, impervious surface, bare soil, \\water, coniferous, deciduous, brushwood, vineyard, \\herbaceous vegetation, agricultural land, plowed land \end{tabular}\\  \hline
FAST~\cite{wang2023samrs} & 37 & \begin{tabular}[c]{@{}l@{}} A220, A321, A330, A350, ARJ21, baseball field, basketball court, \\Boeing737, Boeing747, Boeing777, Boeing787, bridge, bus, C919, cargo truck, \\dry cargo ship, dump truck, engineering ship, excavator, fishing boat, \\football field, intersection, liquid cargo ship, motorboat, other-airplane, \\other-ship, other-vehicle, passenger ship, roundabout, \\small car, tennis court, tractor, trailer, truck tractor, tugboat, van, warship\end{tabular}\\  \hline
Potsdam~\cite{ISPRS_Potsdam} & 6 & \begin{tabular}[c]{@{}l@{}} impervious surface, building, \\low vegetation, tree, car, clutter\end{tabular}\\  \hline
FloodNet~\cite{floodnet} & 9 & \begin{tabular}[c]{@{}l@{}} building-flooded, building-non-flooded, road-flooded, road-non-flooded, \\water, tree, vehicle, pool, grass\end{tabular}\\ 
\bottomrule
\end{tabular}
\end{adjustbox}
\vspace{0.1cm}
\caption{\textbf{Category Names for datasets used in our semantic segmentation benchmarks.} 
}
\label{tab:dataset-category-semseg}
\end{table*}

\end{document}